\documentclass{article} 
\usepackage{iclr2027_conference,times}
\usepackage[T1]{fontenc}

\usepackage{amsmath,amsfonts,bm}

\def\eqref#1{equation~\ref{#1}}

\def\1{\bm{1}}

\DeclareMathAlphabet{\mathsfit}{\encodingdefault}{\sfdefault}{m}{sl}
\SetMathAlphabet{\mathsfit}{bold}{\encodingdefault}{\sfdefault}{bx}{n}

\usepackage{hyperref}
\usepackage{url}
\usepackage{amssymb}
\usepackage{graphicx}
\usepackage{float}
\usepackage{bm} 
\usepackage{colortbl}
\usepackage{array}
\usepackage[nameinlink]{cleveref}
\DeclareRobustCommand{\parenCref}[1]{%
  (\hyperref[#1]{\nameCref{#1}~\ref*{#1}})%
}
\hypersetup{colorlinks, linkcolor={red!65!black}, citecolor={blue!75!black}, urlcolor={blue!75!black}}
\usepackage{cancel}
\usepackage{tcolorbox}
\usepackage{listings}
\usepackage{inconsolata}
\definecolor{codekeyword}{HTML}{008000}
\definecolor{codename}{HTML}{0000FF}
\definecolor{codecomment}{HTML}{3D7B7B}
\definecolor{codestring}{HTML}{BA2121}
\definecolor{codeoperator}{HTML}{666666}
\lstdefinestyle{python}{
    language=Python,
    basicstyle=\fontsize{8.4pt}{10pt}\selectfont\ttfamily,
    keywordstyle=\color{codekeyword}\bfseries,
    keywordstyle=[2]\color{codekeyword},
    deletekeywords=[2]{sum},
    emph={score_centering_loss},
    emphstyle=\color{codename},
    morekeywords=[3]{jax,numpy,lax},
    keywordstyle=[3]\color{codename}\bfseries,
    commentstyle=\color{codecomment}\itshape,
    stringstyle=\color{codestring},
    literate=*
        {0}{{\textcolor{codeoperator}{0}}}1
        {1}{{\textcolor{codeoperator}{1}}}1
        {2}{{\textcolor{codeoperator}{2}}}1
        {3}{{\textcolor{codeoperator}{3}}}1
        {4}{{\textcolor{codeoperator}{4}}}1
        {5}{{\textcolor{codeoperator}{5}}}1
        {6}{{\textcolor{codeoperator}{6}}}1
        {7}{{\textcolor{codeoperator}{7}}}1
        {8}{{\textcolor{codeoperator}{8}}}1
        {9}{{\textcolor{codeoperator}{9}}}1
        {=}{{\textcolor{codeoperator}{=}}}1
        {+}{{\textcolor{codeoperator}{+}}}1
        {-}{{\textcolor{codeoperator}{-}}}1
        {*}{{\textcolor{codeoperator}{*}}}1
        {/}{{\textcolor{codeoperator}{/}}}1,
    backgroundcolor=\color{gray!6},
    frame=single,
    rulecolor=\color{gridline},
    framesep=6pt,
    xleftmargin=6pt,
    xrightmargin=6pt,
    showstringspaces=false,
    columns=fullflexible,
    keepspaces=true,
    upquote=true,
    tabsize=4,
    breaklines=false
}

\newtcolorbox{notationbox}{colback=gridgray, colframe=gridgray, boxrule=0pt, arc=2pt, left=6pt, right=6pt, top=4pt, bottom=4pt, boxsep=0pt}
\definecolor{gentlelilac}{RGB}{112,48,175}

\title{Score Centering Stabilizes \\ Off-policy Reinforcement Learning}

\author{Martin Marek \& Max Ryabinin \\
Together AI \\
\vspace{-8mm}
}

\definecolor{gridblue}{RGB}{235,244,250}
\definecolor{gridcyan}{RGB}{234,247,247}
\definecolor{gridgreen}{RGB}{238,247,238}
\definecolor{gridamber}{RGB}{252,246,229}
\definecolor{gridorange}{RGB}{252,240,229}
\definecolor{gridyellow}{RGB}{250,248,225}
\definecolor{gridred}{RGB}{252,238,238}
\definecolor{gridteal}{RGB}{232,247,241}
\definecolor{gridpurple}{RGB}{246,239,250}
\definecolor{gridpink}{RGB}{251,237,246}
\definecolor{gridgray}{RGB}{243,243,243}
\definecolor{deep0}{HTML}{4C72B0}      
\definecolor{deep1}{HTML}{DD8452}
\definecolor{deep2}{HTML}{55A868}
\definecolor{tblteal}{HTML}{2A9D8F}   
\definecolor{figother}{HTML}{666666}  
\definecolor{gridline}{RGB}{210,210,210}
\definecolor{driftcoral}{HTML}{C65D4A}
\definecolor{signalteal}{HTML}{087F78}
\definecolor{equationnavy}{HTML}{283348}

\iclrfinalcopy 
\begin{document}

\vspace*{-5mm} 
\maketitle

\begin{abstract}
Reinforcement learning (RL) of large language models is notoriously sensitive to small differences between training and inference engines, often referred to as the training-inference mismatch (TIM).
However, completely eliminating TIM is impractical, as it would come at a major cost to rollout efficiency.
In this paper, we show that the instability of RL under TIM is primarily caused by \textit{drift}: a persistent bias between training and inference engines that accumulates with every training step.
We derive an additive \textit{``score centering''} correction term that stabilizes RL under TIM by canceling drift.
When training models from 0.6B to 30B parameters, score centering alone matches or outperforms methods based on importance sampling under quantization, with the gap growing as the mismatch becomes more severe.
Because the correction is additive, score centering also composes with importance sampling -- their composition outperforms pure importance sampling baselines in our staleness experiments.
\end{abstract}

\begin{figure}[H]
    \centering
    \includegraphics[width=1\linewidth]{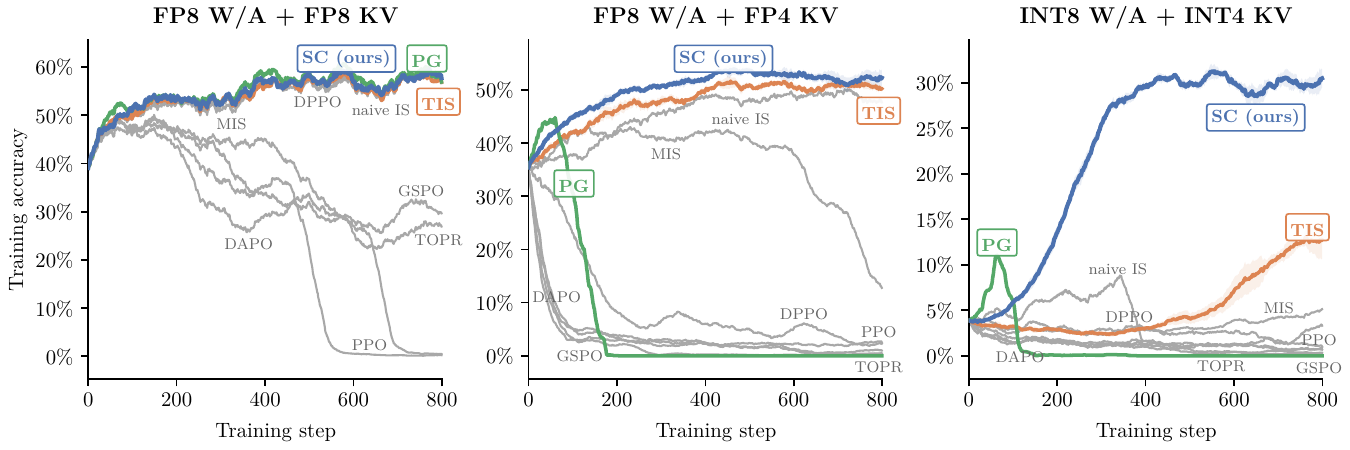}
    \caption{
    \textbf{Score centering outperforms importance sampling under heavy quantization.}
    We train Qwen3-30B-A3B-Base~\citep{qwen3} on INTELLECT-2 math data~\citep{intellect2} using RL under progressively stronger quantization of the sampler.
    We compare correction rules on a shared objective (\Cref{sec:results_setup}).
    With mild quantization, naive \textcolor{deep2}{policy gradient (PG)} trains stably over 800 steps, and performs on par with correction methods.
    As quantization severity increases, training becomes less stable, and the gap between \textcolor{deep0}{score centering (SC)}, \textcolor{deep1}{truncated importance sampling (TIS)}, and \textcolor{figother}{other correction methods} widens.
    }
    \vspace{12pt}
    \label{fig:30b_math}
\end{figure}

\section{Introduction}

Reinforcement learning is a crucial step in the training of today's large language models, as reflected by an increasingly large fraction of training compute allocated towards it~\citep{o1,deepseek_r1,scaleRL,mai_1}. 
This is typically achieved through algorithms such as PPO, GRPO, DAPO, and IcePop~\citep{ppo,grpo,dapo,icepop}, which are all based on the policy gradient method~\citep{reinforce}.

RL training of LLMs via policy gradient consists of sampling rollouts, assigning a reward to each rollout, and computing gradients on the weighted rollouts. 
More formally, the policy gradient method computes the gradient of the expected reward over rollouts $y$ from the policy $p_\theta$ as the expected score weighted by the reward $R$:
\begin{equation}
\nabla_\theta \, \mathbb{E}_{p_\theta}[R] = \mathbb{E}_{p_\theta} \big[ R \, \underbrace{\nabla_\theta \log p_\theta(y)}_{\text{``score''}} \big].
\label{eq:policy_grad}
\end{equation}

Crucially for practitioners, computing the right side of \Cref{eq:policy_grad} requires two separate forward passes through the model: one forward pass of the sampler (inference engine) to generate the rollouts and a second forward pass of the trainer to compute gradients on the generated rollouts. 
In theory, the trainer and sampler represent the same model, so their outputs should be identical. 
However, in practice, this is rarely the case. 
Most RL codebases admit small numerical differences between training and inference engines, resulting in the \textit{training-inference mismatch} (TIM), which can degrade training performance \citep{scaleRL} or even lead to complete reward collapse \citep{flashrl}. By contrast, supervised fine-tuning generally has stable training dynamics using samples from a different model or a human annotator, with no adjustments required.

In this work, we investigate the reasons behind the sensitivity of RL to the training-inference mismatch. 
By analyzing the policy gradient update, we show that it has a \textit{drift} term, equal to 0 if and only if training and sampling policies match.
Based on this finding, we introduce a novel method to stabilize RL training under TIM, called \textbf{score centering}.\footnote{Code: \href{https://github.com/martin-marek/score-centering}{\texttt{github.com/martin-marek/score-centering}}}
As we show in experiments with Qwen3-0.6B on Countdown and Qwen3-30B-A3B-Base on the math subset of INTELLECT-2 data, score centering can match or outperform state-of-the-art techniques for training stabilization based on importance sampling.
Due to the orthogonality of the two approaches, score centering can be combined with importance sampling for further gains.

\section{Background and Related Work}
\label{sec:background}

\subsection{Sources of Training-Inference Mismatch}

The vanilla policy gradient identity in \Cref{eq:policy_grad} only holds when the policy for computing the gradients is the same policy that produced the rollouts.
In practice, however, rollouts are generated from a \textbf{sampler $\bm{q_\theta}$}, while gradients are computed through a \textbf{trainer $\bm{p_\theta}$}. 
Hence, training-inference mismatch occurs when $q_\theta \ne p_\theta$.

There are many sources of TIM, each with different severity. In the worst case, the training and inference engines run at different precisions or are based on two independent codebases, each of which can have different (hard-to-detect) bugs that result in subtly different outputs \citep{speed_kills_stability, flashinfer_sentinel}. 
In a less severe case, even if both engines are correct and use the same GPU kernels, they might still produce different outputs due to non-associativity of floating point operations.\footnote{For example, in \texttt{fp4\_e2m1} precision, $(0.5+0.5)+2=3$, but $0.5+(0.5+2)=2$.}
Since sampling is autoregressive, while training is typically parallelized over the sequence dimension, the training and inference engines might call the same kernels with different input shapes, changing the order of reductions, resulting in small floating point differences. 
Batch-invariant kernels eliminate the dependency on input shapes, but require extensive engineering work and result in worse GPU utilization \citep{thinky_nondeterminism}. 
Switching both engines from bf16 to fp16 precision also greatly reduces the mismatch \citep{fp16_rl}, but does not eliminate it entirely.

Finally, even if the trainer and the generator use identical batch-invariant kernels, TIM can occur due to update staleness. 
Maximizing GPU utilization during RL requires disaggregating training and inference engines, running them asynchronously \citep{areal}, and serving rollouts with continuous batching \citep{orca}. 
This means that a single training batch, or even different subsequences of a single rollout, can be generated from different model checkpoints \citep{pipelineRL}. 
Staleness becomes most severe in long-context environments such as agentic coding, where some episodes may finish in minutes, while others might last hours or even days \citep{single_rollout}. 
From the perspective of hardware utilization, it is highly desirable to algorithmically stabilize training with stale rollouts, rather than trying to eliminate staleness entirely \citep{areal}.

Our goal is to propose a method for stable RL training under TIM, to enable high hardware utilization while ensuring stable training.

\subsection{Importance-Sampling Corrections}

Assume that in \Cref{eq:policy_grad}, the rollouts come from the sampler $q_\theta$, rather than the trainer $p_\theta$, where $q_\theta \ne p_\theta$. Then we can correct for the training-inference mismatch exactly using importance sampling (IS):
\begin{equation}
\nabla_\theta \, \mathbb{E}_{p_\theta}[R] = \mathbb{E}_{p_\theta} \left[ R \, \nabla_\theta \log p_\theta(y) \right] = \mathbb{E}_{q_\theta} \left[ \frac{p_\theta(y)}{q_\theta(y)} R \, \nabla_\theta \log p_\theta(y) \right].
\label{eq:policy_grad_is}
\end{equation}
\Cref{eq:policy_grad_is} is written for full sequences; in practice, the ratio is typically applied per token \citep{gspo}. Either way, importance sampling corrects for TIM at the cost of increased variance. The weighted score gets multiplied by the importance ratio $r = \frac{p_\theta(y)}{q_\theta(y)}$, which can take arbitrarily large values on rare tokens, inflating the variance of the gradient estimate \citep{ionides, owen_mc}. Since training with raw importance ratios is unstable, the methods used in practice bound the importance ratio, which inevitably introduces bias.

\Cref{tab:is_methods} (in \Cref{sec:is_methods}) shows that most popular correction methods are based on importance sampling and fall onto a simple grid, depending on the region where the importance ratio gets clipped or masked. A notable outlier is DPPO, which uses a masking region based on binary total variation -- but its objective still uses an IS ratio, just like every other method in the grid. Going beyond the table, \citet{alp} inject learnable perturbations into the trainer's hidden states and use the perturbed policy as the numerator of the importance ratio -- reducing the heavy tail of the ratios instead of clipping them -- but still relying on importance sampling. In contrast, \textit{score centering} (a method that we introduce in \Cref{sec:sc}) uses \textit{no} importance ratios, no masking, and no clipping -- it is an additive correction term that works fundamentally differently from every method in this grid.

\section{Why RL Is Sensitive to Mismatch}
\label{sec:diagnosis}

To understand why RL is so sensitive to TIM, it helps to understand what makes RL different from Supervised Fine-Tuning (SFT). 
As we noted earlier, SFT is stable under a far more severe mismatch -- training on data generated completely offline by a different model, with no importance sampling correction~\citep{distillation, deepseek_r1}. 
In other words, SFT is not sensitive to staleness or the choice of kernels at all.

Crucially, policy gradient reduces to SFT on a task with binary rewards when the advantages are $+1/0$ (i.e.\ the rewards are not centered or normalized) and the training is completely offline (i.e.\ the sampler never gets updated).\footnote{In practice, advantages almost always include negative values (for example, because of group centering \citep{grpo}), and even if training admits some level of staleness, it is never completely offline.} Therefore, we can consider the \textbf{reward mode} and \textbf{staleness} to be the two key differences between SFT and RL.\footnote{Note that shifting the rewards by a constant does not change the expected gradient on-policy, and group centering only rescales it by $1 - 1/G$ for a group of $G$ rollouts -- the reason to center rewards is that it reduces the gradient variance \citep{greensmith, dr_grpo}. 
However, this doesn't hold off-policy: SFT with $+1/0$ converges to a maximum likelihood estimator \citep{bishop}, while SFT with $0/-1$ rewards is degenerate: the loss becomes unbounded and can result in model collapse \citep{finetuning_dynamics}.} 
We want to determine which of these two mechanisms is the one that results in increased sensitivity to TIM.

We test this using a toy setup of training with intentional policy mismatch. 
We train Qwen3-1.7B \citep{qwen3} on Countdown \citep{stream_of_search}, and artificially perturb the weights of the sampler by a small amount compared to the trainer, to create controlled TIM. We provide further justification behind this setup in \Cref{sec:results} as well as test more realistic sources of TIM.


\begin{figure}[H]
    \centering
    \includegraphics[width=0.9\linewidth]{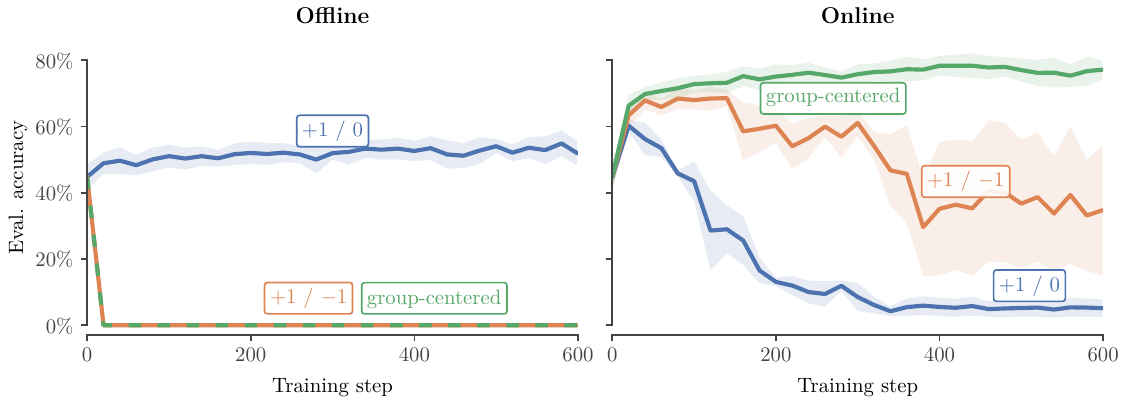}
    \caption{Offline vs.\ online training of Qwen3-1.7B on Countdown under three reward modes. Offline training is \emph{only} stable with $+1/0$ rewards, while online training under TIM is \emph{least} stable with $+1/0$ rewards.}
    \label{fig:2b_online_vs_offline}
\end{figure}

\Cref{fig:2b_online_vs_offline} shows that in this setup, offline training is only stable with $+1/0$ (non-negative) rewards, while the opposite holds for online training -- group-centered rewards \citep{grpo}, which mix positive and negative values within a group, are the \textit{most} stable, while $+1/0$ rewards -- that were most stable for offline training -- are \textit{least} stable for online training. We therefore hypothesize that there are two separate mechanisms at play: one making offline training with \textit{negative} rewards unstable, and another one making online training with \textit{positive} rewards unstable.

The main focus of this paper is online reinforcement learning, therefore we do not explore the instability of offline training with negative rewards further. \Citet{finetuning_dynamics} argue that the main mechanism behind this instability is the unboundedness of negative rewards in combination with distribution sharpening. Logprobs are unbounded from below, therefore during offline training with negative rewards, they can diverge to $-\infty$. This effect is absent from online training where only high probability tokens get sampled. 
This asymmetry is well documented: training on positive samples alone (i.e.\ distillation or rejection finetuning) is completely stable even offline \citep{distillation, deepseek_r1}, while negative gradients can drag down the probability of correct responses \citep{nthr} -- even though on-policy, they carry useful learning signal \citep{negative_reinforcement}.

In online training, we observe the opposite behavior to offline training -- using only positive rewards is the \emph{least} stable setting. We hypothesize therefore that the instability must arise from the online nature of the training process. 
Note that online training with $+1/0$ rewards under TIM is itself a form of distillation: the trainer is fit to successful rollouts from the sampler, a biased copy of itself, whose weights are refreshed from the trainer after every step. The only difference from offline distillation is that the teacher moves with the student, so this is where the instability must come from.

\section{Score Centering}
\label{sec:sc}

\begin{notationbox}
\textbf{Notation.} The update from a rollout $y$ sums its token scores, each weighted by the rollout's reward (or advantage) $R$. We examine the expected contribution of one token $y_t$, conditional on its prefix $y_{<t}$. The sampler $q$ produces the token $y_t$; $v$ indexes the vocabulary. $p_v$ and $q_v$ are the next-token probabilities of $v$ under the trainer $p$ and the sampler $q$, $s_v = \nabla_\theta \log p_v$ is its score, and $\bar{s} = \sum_v q_v \, s_v$ is the expected score under the sampler. $\mathbb{E}_q$, $\mathbb{E}_p$ are over the rest of the rollout after the prefix.
\end{notationbox}

Consider RL training with vanilla policy gradient in an environment with a constant $+1$ reward. In theory, there should be no learning -- the reward is constant, so there is no ``signal'' coming from the environment. And indeed, on-policy the expected gradient is zero at every prefix: $\mathbb{E}_p[R \, s_{y_t}] = R \, \mathbb{E}_p[s_{y_t}] = R \cdot 0$, since $\mathbb{E}_p[s_{y_t}] = \sum_v p_v \nabla_\theta \log p_v = \nabla_\theta \sum_v p_v = \nabla_\theta 1 = 0$ for any distribution.

\clearpage
Under TIM, however, the expected gradient is nonzero: the token is sampled from the sampler $q$, while the score is computed through the trainer $p$, so $\bar{s} = \mathbb{E}_q[s_{y_t}] \ne 0$ in general. \begingroup\color{equationnavy}
Using the covariance identity,\footnote{$\mathrm{Cov}(A, B) = \mathbb{E}[AB] - \mathbb{E}[A]\,\mathbb{E}[B] \; \Rightarrow \; \mathbb{E}[AB] = \mathbb{E}[A]\,\mathbb{E}[B] + \mathrm{Cov}(A, B)$} we can decompose the expected policy gradient update at a prefix into a \textcolor{driftcoral}{``drift''} and a \textcolor{signalteal}{``signal''} term:
\begin{equation}
\underbrace{\mathbb{E}_q[R \, s_{y_t}]}_{\text{policy gradient update}}
\hspace{-3mm} = \hspace{1.3mm} \textcolor{driftcoral}{\underbrace{\mathbb{E}_q[R] \; \bar{s}}_{\text{``drift''}}}
\; + \; \textcolor{signalteal}{\underbrace{\mathrm{Cov}_q(R, \; s_{y_t})}_{\text{``signal''}}}
\label{eq:drift}
\end{equation}
\endgroup
\vspace{-1mm}
\paragraph{Drift is an artifact of TIM that acts as distillation toward the sampler.} The drift term carries no information about which rollouts were successful: it depends on the rewards only through their mean $\mathbb{E}_q[R]$. Only the covariance term sees which token led to which reward. What the drift term does instead is determined by its direction $\bar{s}$: the expected score is the negative gradient of the cross-entropy (SFT) loss with the sampler as the teacher, so vanilla policy gradient distills the trainer toward the sampler at every prefix, scaled by the expected reward at that prefix. This is purely an artifact of TIM: on-policy, the trainer already matches the sampler, so $\bar{s}=0$ and the drift term vanishes. Distillation toward a \textit{fixed} teacher is harmless -- it converges to the maximum likelihood fit of that teacher. The sampler, however, is not fixed: it is a biased copy of the trainer (e.g.\ quantized or stale), so each step pushes the trainer toward the sampler; the trainer's weights are then synced back to the sampler, and the error compounds in a feedback loop instead of converging. This explains why online training with all-positive rewards is the least stable setting in \Cref{fig:2b_online_vs_offline}. \citet{probing} likewise identify systematic bias as the root cause of instability under TIM, and show it compounds through a positive feedback loop.

\paragraph{Group centering does not remove drift.} Group centering \citep{grpo} makes the advantages sum to zero over the rollouts of a prompt, but drift arises at individual prefixes, and the expected advantage at a given prefix is not zero: a prefix that is likely to lead to a correct answer has positive expected advantage, and a prefix that already contains a mistake has negative expected advantage. Drift is therefore nonzero precisely at the prefixes that carry learning signal -- the trainer is pulled toward the sampler after prefixes with positive expected advantage and pushed away from it after prefixes with negative expected advantage, regardless of whether the tokens in question affect the reward. Group centering does shrink drift relative to $+1/0$ rewards, which is consistent with it being the most stable online setting in \Cref{fig:2b_online_vs_offline}, but it does not eliminate it.

\paragraph{Score centering removes drift.} These observations directly motivate our method, called \emph{score centering}: even under TIM, we want the expected score to be zero at every prefix, as it is on-policy. We achieve this simply by subtracting from each score the expected score under the sampler, replacing $s_{y_t}$ by the centered score $\tilde{s}_{y_t}$:
\begin{equation}
\tilde{s}_{y_t} = s_{y_t} - \bar{s}.
\label{eq:sc}
\end{equation}
The expected policy gradient update with score centering then becomes:
\begin{align}
\hspace{25mm} \mathbb{E}_q[R\,\tilde{s}_{y_t}] &= \overset{\color{gentlelilac} \mathbb{E}_q[\tilde{s}_{y_t}] \,=\, \bar{s} - \bar{s} \,=\, 0}{\cancel{\mathbb{E}_q[R]\;\mathbb{E}_q[\tilde{s}_{y_t}]}} + \mathrm{Cov}_q(R, \tilde{s}_{y_t}) \label{eq:sc_bias} \\
&= \mathrm{Cov}_q(R, \tilde{s}_{y_t}) \nonumber \\
&= \mathrm{Cov}_q(R, s_{y_t}) \;\; \longleftarrow \;\;
\smash[t]{\begin{array}{@{}l@{}}
\text{only difference to on-policy update} \\
\text{is the covariance sampling distribution}
\end{array}} \nonumber
\end{align}

The centered score as defined in \Cref{eq:sc} has zero mean under the sampler because both expectations are taken under the same distribution -- the drift gets canceled exactly at every prefix, even off-policy. \Cref{eq:sc_bias} shows that the expected update with score centering equals the expected on-policy update up to the distribution over which the covariance is measured -- both on-policy and under TIM with importance sampling, the covariance is measured under the training distribution $p_\theta$, while score centering measures it under the sampling distribution $q_\theta$. In the absence of TIM, i.e.\ when $p_\theta = q_\theta$, score centering is a no-op, since the correction term in \Cref{eq:sc} is exactly zero.

\paragraph{Score Centering vs Importance Sampling.} Importance sampling, just like score centering, also cancels drift exactly. However, importance sampling uses a random multiplicative correction term, meaning that it increases the gradient variance, and its value depends on the sampled token. For rare tokens, the importance ratio can be very large, which is why almost all practical implementations clip or mask large importance ratios in some way (\Cref{tab:is_methods}), reintroducing drift. In contrast, score centering is an \textit{additive} correction term that is \textit{deterministic} given the prefix -- it does not depend on the sampled token, and it can be evaluated exactly. Since score centering and importance sampling correct for TIM in independent ways, the two methods can be composed with each other, as described in \Cref{sec:is_composition}.

\paragraph{Relation to classical baselines.} Subtracting a zero-mean quantity from the policy gradient is a classical variance-reduction idea: reward baselines \citep{reinforce} and score-function control variates \citep{bbvi} both rely on the identity $\mathbb{E}_p[s_{y_t}] = 0$. Score centering is a bias correction rather than a variance reduction: the subtracted vector is deterministic given the prefix and is chosen to shift the mean of the update, whereas on-policy a reward baseline changes the variance but not the mean. Off-policy, a reward baseline equal to the expected reward at the prefix (an exact value function) would also cancel drift in expectation; score centering obtains the same expected update without a critic. To our knowledge, no prior method subtracts the expected score itself: on-policy it is zero, and in classical off-policy RL it requires an intractable expectation over the action space, so off-policy methods rely on importance sampling instead. RL training of LLMs under TIM is unusual in that the expected score is both nonzero and computable exactly, as a sum over the next token's logprobs.

\subsection{Implementation}
\label{sec:sc_impl}

Storing the sampler's full next-token distribution is prohibitively expensive. We therefore log only its top-$k$ logprobs ($k=128$) and model the tail with the trainer's distribution, rescaled to match the sampler's tail mass. We implement score centering as a scalar loss:
\begin{equation}
L = -R \, \Big( \log p_{y_t} - \sum_{v \in H} \operatorname{sg}\left[ q_v - \rho \, p_v \right] \log p_v \Big),
\label{eq:sc_loss_main}
\end{equation}
where $\rho$ is the ratio of the sampler's to the trainer's tail mass and $\operatorname{sg}$ denotes stop-gradient. Using $k=128$ or even $k=32$ matches full score centering in every setting we tested (\Cref{sec:topk_ablation}). The code snippet below illustrates a minimal implementation for a single token:
\begin{lstlisting}[style=python]
import jax.numpy as jnp
from jax.lax import stop_gradient

def score_centering_loss(train_logp, samp_logp, topk_ids, sampled_token, advantage):
    train_head_logp = train_logp[topk_ids]
    tail_mass_ratio = (1 - jnp.exp(samp_logp).sum()) / (1 - jnp.exp(train_head_logp).sum())
    head_prob_residual = jnp.exp(samp_logp) - tail_mass_ratio * jnp.exp(train_head_logp)
    logp_correction = (stop_gradient(head_prob_residual) * train_head_logp).sum()
    return -advantage * (train_logp[sampled_token] - logp_correction)
\end{lstlisting}

In \Cref{sec:sc_implementation}, we provide the full derivation and a generalized implementation of score centering with support for importance weights.

\section{Experiments}
\label{sec:results}

\subsection{Setup}
\label{sec:results_setup}

We compare score centering against the correction methods in \Cref{tab:is_methods} by training Qwen3-0.6B-Instruct on Countdown \citep{stream_of_search} and Qwen3-30B-A3B-Base on the math subset of the {INTELLECT-2} dataset \citep{intellect2}.

\noindent\textbf{Shared objective.} Existing methods often bundle several techniques together: for example, DAPO combines asymmetric clipping with dynamic sampling \citep{dapo}, while IcePop combines MIS between training and inference engines with PPO clipping between policy versions \citep{icepop}. To compare each correction method in isolation, we use the same objective (REINFORCE with group-centered rewards) throughout, with each correction being applied on top of this objective in isolation. All methods share the same sampler, trainer, optimizer and advantages, with one SGD step per batch. All importance ratios use the sampler's logged probabilities. \Cref{tab:is_methods} shows the correction rules and hyperparameters, together with implementation notes. The learning rate, batch sizes, and compute costs are listed in \Cref{sec:experimental_details}.

\noindent\textbf{Mismatch settings.} In general, we struggled to find natural setups with TIM severe enough to observe statistically significant differences between the strongest-performing methods within a handful of GPU hours.
We believe that this is a consequence of the accumulation of drift during training (\Cref{sec:sc}): under mild TIM, it takes more training steps for sufficient drift to accumulate, and until it does, the best-performing methods overlap.
We show this in \Cref{sec:results_weight_noise}: under small weight noise, the best-performing methods overlap, making it impossible to draw comparisons, while as TIM increases, methods collapse earlier and the performance gap between them grows. Since we cannot afford to ablate correction methods with statistical significance over very long training runs, we deliberately amplify TIM instead, and use these results as a proxy for large-scale training under milder TIM. For this reason, in \Cref{sec:results_quant_stale} we test intentionally severe quantization and staleness settings. Notably, prior works disagree on just how unstable RL under TIM really is: \citet{fp8_rl} find fp8 rollouts stable using only token-level importance sampling without matched numerics, while \citet{mai_1} see runs diverge despite trying to minimize TIM and using bf16 sampling. We believe both findings are consistent with drift accumulating over training: to observe instability within short training runs, TIM must be severe, while mild TIM can destabilize training too, but it takes more training steps.

\subsection{Synthetic Weight Noise}
\label{sec:results_weight_noise}

For fast initial experimentation, we found it useful to study TIM induced by adding an artificial offset to the weights of the sampler compared to the weights of the trainer. Namely, we set $\theta_\textrm{sampler} = \theta_\textrm{trainer} \, + \, \Delta \theta$, where $\Delta \theta$ was sampled at the beginning of training from an isotropic Gaussian distribution and held fixed during training. 
While this is the least realistic setup we have studied, it resulted in the fastest collapse and separation between methods, and allowed for a continuous scale of TIM, making it an indispensable tool for initial experiments.

\begin{figure}[H]
    \centering
    \includegraphics[width=\linewidth]{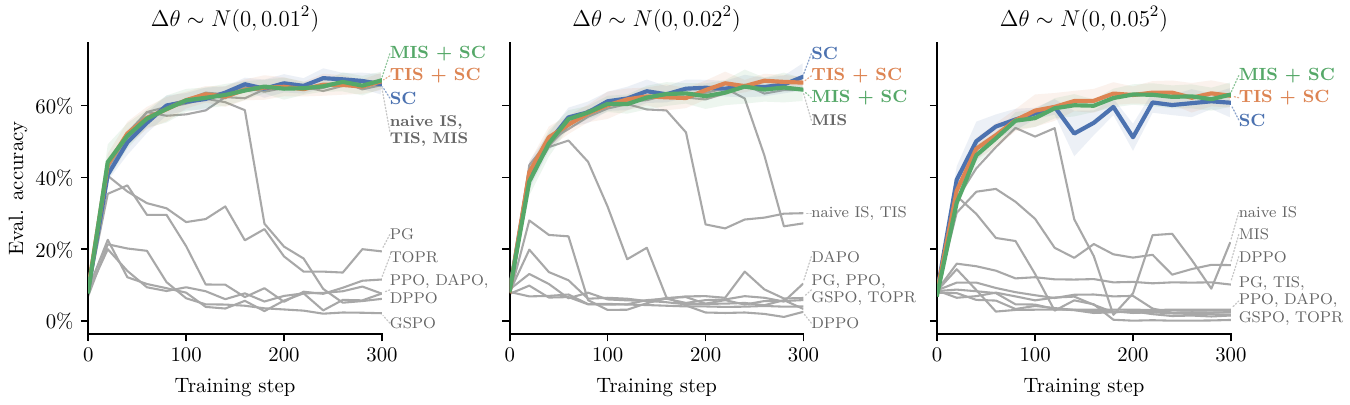}
    \caption{Qwen3-0.6B on Countdown with Gaussian noise added to the sampler weights. More noise causes earlier collapse. Only score centering (alone or composed with TIS / MIS) trains stably under the largest noise.}
    \label{fig:06b_weight_noise}
\end{figure}

Across three noise scales in \Cref{fig:06b_weight_noise}, we find Score Centering (SC), Truncated Importance Sampling (TIS) and Masked Importance Sampling (MIS) to perform the best, along with their compositions: MIS + SC and TIS + SC.
We describe in \Cref{sec:is_composition} how score centering can be composed on top of importance sampling methods. Under the most severe noise, only score centering and score centering composed with MIS or TIS trained stably. \Cref{fig:06b_weight_noise} also shows drift accumulating over training -- methods that collapse do so earlier under larger mismatch. For example, DPPO collapses at roughly step 160, 80 and 20 across the three noise scales, and TIS trains stably under the smallest noise but collapses at roughly step 180 and 40 under the larger two.

\subsection{Quantization and Staleness}
\label{sec:results_quant_stale}

Next, we replace artificial weight noise with two more realistic sources of TIM: quantization and staleness. 
We deliberately made both sources of TIM severe (\Cref{sec:results_setup}) to observe the separation between different correction methods within hundreds of training steps using short sequence lengths.
In the quantized setting, we quantize only the sampler (weights, activations, and KV cache); the trainer uses bf16 precision. The staleness setting is intentionally severe too: the inference engine gets updated \textit{only} every 64 steps, rather than having a \textit{maximum} staleness of 64. 
In practice, it would be preferable to update the inference engine as frequently as possible \citep{pipelineRL} and to apply the same quantization scheme to both the sampler and the trainer \citep{jet_rl}.

\begin{figure}[H]
    \centering
    \includegraphics[width=\linewidth]{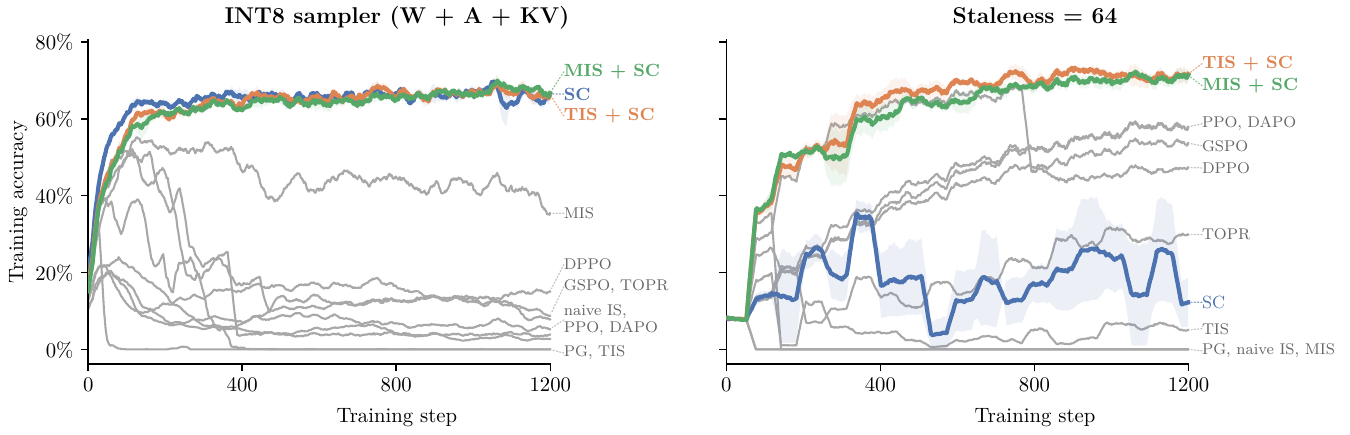}
    \caption{Qwen3-0.6B on Countdown with an int8 sampler (left) and a sampler updated only every 64 steps (right). Under quantization, both score centering alone and composed perform best; under staleness, score centering composed with TIS / MIS dominates.}
    \label{fig:06b_quant_stale}
\end{figure}

In the quantized sampler setting (\Cref{fig:06b_quant_stale}), we see similar results to \Cref{fig:06b_weight_noise} -- score centering, alone and composed on top of TIS / MIS, performs best. However, under the staleness setting, score centering composed with TIS / MIS performs best, outperforming vanilla score centering. We attribute this to score centering measuring the covariance under the sampler rather than the trainer \parenCref{eq:sc_bias}: TIS partially corrects the sampling distribution, and score centering removes the remaining drift. We therefore recommend composing score centering with an importance-sampling correction whenever the trainer can move far from the sampler between syncs. PPO and DAPO survive staleness but collapse under quantization and weight noise. This is consistent with their clip being designed for ratios that arise from policy movement: staleness produces such ratios, while numerical error does not.

\subsection{Scaling to 30B}
\label{sec:results_30b}

Finally, we verify that our results hold at a larger scale by training Qwen3-30B-A3B-Base on INTELLECT-2 math under three levels of sampler quantization (\Cref{fig:30b_math}). Since Qwen3-30B-A3B is a mixture-of-experts model, the sampler and the trainer can also disagree on expert routing; we do not replay router indices, although doing so would be preferable in practice \citep{r3}. With an FP8 sampler, even uncorrected policy gradient trains stably, reaching $58\%$ training accuracy. Once the KV cache is quantized to FP4, policy gradient collapses within 200 steps and MIS collapses late in training, while score centering ($52\%$) and TIS ($51\%$) stay stable. With an INT8 sampler and an INT4 KV cache, score centering reaches $30\%$, TIS $12\%$, and every other method ends below $5\%$.


\section{Conclusion}

This paper makes two contributions. The first is an explanation of why RL training of LLMs is so sensitive to training-inference mismatch. Under TIM, the policy gradient update contains a drift term that acts as distillation toward the sampler. Because the sampler is a biased copy of the trainer and is periodically synced back to it, this bias compounds in a feedback loop: the more severe the TIM, the earlier training collapses. Prior work has observed that this bias compounds over training \citep{probing, mai_1}. We show that canceling drift alone, with no importance ratios, is sufficient to stabilize training under severe quantization, which establishes drift as the cause of the instability. This also explains why offline distillation from a completely different model is stable, while online RL collapses under small numerical differences: what matters is not the size of the mismatch, but whether the teacher is fixed or tracks the student.

Our second contribution is score centering, a practical method that cancels drift by subtracting the expected score. It is an additive correction term with no hyperparameters, it can be expressed as a scalar loss \parenCref{eq:sc_loss_main}, and it composes with importance-sampling methods such as TIS and MIS. Exact score centering requires the sampler's full next-token distribution; we therefore introduce an approximation that needs only the sampler's top-$k$ logprobs and matches the performance of exact score centering in all of our experiments (\Cref{sec:topk_ablation}). Under mild TIM, score centering matches importance-sampling methods; under severe quantization, it is the only method that trains stably, and under severe staleness its composition with TIS or MIS performs best.

\paragraph{Limitations.} Score centering cancels drift, but the remaining update measures the covariance between reward and score under the sampler rather than the trainer \parenCref{eq:sc_bias}. This mismatch matters under severe staleness, where score centering performs best when composed with importance sampling (\Cref{sec:results_quant_stale}). Additionally, our headline results come from deliberately severe mismatch on short sequences, used as a proxy for long training under milder mismatch.

\bibliography{paper.bib}

@online{speed_kills_stability,
  title = {When Speed Kills Stability: Demystifying {RL} Collapse from the Training-Inference Mismatch},
  author = {Liu, Jiacai and Li, Yingru and Fu, Yuqian and Wang, Jiawei and Liu, Qian and Jiang, Zhuo},
  year = {2025},
  month = sep,
  url = {https://richardli.xyz/rl-collapse}
}

@article{thinky_nondeterminism,
  author = {Horace He and {Thinking Machines Lab}},
  title = {Defeating Nondeterminism in {LLM} Inference},
  journal = {Thinking Machines Lab: Connectionism},
  year = {2025},
  note = {https://thinkingmachines.ai/blog/defeating-nondeterminism-in-llm-inference/},
  doi = {10.64434/tml.20250910}
}

@article{pipelineRL,
  title={{PipelineRL}: Faster On-Policy Reinforcement Learning for Long Sequence Generation},
  author={Pich{\'e}, Alexandre and Kamalloo, Ehsan and Pardinas, Rafael and Chen, Xiaoyin and Bahdanau, Dzmitry},
  journal={arXiv preprint arXiv:2509.19128},
  year={2025}
}

@article{verl,
  author  = {Sheng, Guangming and Zhang, Chi and Ye, Zilingfeng and Wu, Xibin and Zhang, Wang and Zhang, Ru and Peng, Yanghua and Lin, Haibin and Wu, Chuan},
  title   = {{HybridFlow}: A Flexible and Efficient {RLHF} Framework},
  journal = {arXiv preprint arXiv:2409.19256},
  year    = {2024},
  url     = {https://arxiv.org/abs/2409.19256}
}

@techreport{mai_1,
  author = {{Microsoft AI Team}},
  title  = {{MAI-Thinking-1}: Building a Hill-Climbing Machine},
  year   = {2026},
  url    = {https://microsoft.ai/pdf/mai-thinking-1.pdf}
}

@article{reinforce,
  author  = {Williams, Ronald J.},
  title   = {Simple Statistical Gradient-Following Algorithms for Connectionist Reinforcement Learning},
  journal = {Machine Learning},
  volume  = {8},
  pages   = {229--256},
  year    = {1992},
  doi     = {10.1007/BF00992696}
}

@inproceedings{tis,
  author    = {Yao, Feng and Liu, Liyuan and Zhang, Dinghuai and Dong, Chengyu and Shang, Jingbo and Gao, Jianfeng},
  title     = {On the Rollout-Training Mismatch in Modern {RL} Systems},
  booktitle = {NeurIPS 2025 Workshop on Efficient Reasoning},
  year      = {2025},
  url       = {https://openreview.net/forum?id=8MHqvb4lK9},
  note      = {Blog version: \url{https://fengyao.notion.site/off-policy-rl}}
}

@article{cispo,
  author  = {{MiniMax}},
  title   = {{MiniMax-M1}: Scaling Test-Time Compute Efficiently with Lightning Attention},
  journal = {arXiv preprint arXiv:2506.13585},
  year    = {2025},
  url     = {https://arxiv.org/abs/2506.13585}
}

@inproceedings{scaleRL,
  author  = {Khatri, Devvrit and Madaan, Lovish and Tiwari, Rishabh and Bansal, Rachit and Duvvuri, Sai Surya and Zaheer, Manzil and Dhillon, Inderjit S. and Brandfonbrener, David and Agarwal, Rishabh},
  title   = {The Art of Scaling Reinforcement Learning Compute for {LLM}s},
booktitle={The Fourteenth International Conference on Learning Representations},
year={2026},
url={https://openreview.net/forum?id=FMjeC9Msws},
}

@article{icepop,
  author  = {{Ling Team}},
  title   = {Every Step Evolves: Scaling Reinforcement Learning for Trillion-Scale Thinking Model},
  journal = {arXiv preprint arXiv:2510.18855},
  year    = {2025},
  url     = {https://arxiv.org/abs/2510.18855}
}

@article{ppo,
  author  = {Schulman, John and Wolski, Filip and Dhariwal, Prafulla and Radford, Alec and Klimov, Oleg},
  title   = {Proximal Policy Optimization Algorithms},
  journal = {arXiv preprint arXiv:1707.06347},
  year    = {2017},
  url     = {https://arxiv.org/abs/1707.06347}
}

@article{dapo,
  author  = {Yu, Qiying and Zhang, Zheng and Zhu, Ruofei and others},
  title   = {{DAPO}: An Open-Source {LLM} Reinforcement Learning System at Scale},
  journal = {arXiv preprint arXiv:2503.14476},
  year    = {2025},
  url     = {https://arxiv.org/abs/2503.14476}
}

@article{gspo,
  author  = {Zheng, Chujie and Liu, Shixuan and Li, Mingze and Chen, Xiong-Hui and Yu, Bowen and Gao, Chang and Dang, Kai and Liu, Yuqiong and Men, Rui and Yang, An and Zhou, Jingren and Lin, Junyang},
  title   = {Group Sequence Policy Optimization},
  journal = {arXiv preprint arXiv:2507.18071},
  year    = {2025},
  url     = {https://arxiv.org/abs/2507.18071}
}

@article{topr,
  author  = {Le Roux, Nicolas and Bellemare, Marc G. and Lebensold, Jonathan and Bergeron, Arnaud and Greaves, Joshua and Fr\'{e}chette, Alex and Pelletier, Carolyne and Thibodeau-Laufer, Eric and Toth, S\'{a}ndor and Work, Sam},
  title   = {Tapered Off-Policy {REINFORCE}: Stable and Efficient Reinforcement Learning for {LLM}s},
  journal = {arXiv preprint arXiv:2503.14286},
  year    = {2025},
  url     = {https://arxiv.org/abs/2503.14286}
}

@article{dppo,
  author  = {Qi, Penghui and Zhou, Xiangxin and Liu, Zichen and Pang, Tianyu and Du, Chao and Lin, Min and Lee, Wee Sun},
  title   = {Rethinking the Trust Region in {LLM} Reinforcement Learning},
  journal = {arXiv preprint arXiv:2602.04879},
  year    = {2026},
  url     = {https://arxiv.org/abs/2602.04879}
}

@inproceedings{finetuning_dynamics,
  title={Learning Dynamics of {LLM} Finetuning},
  author={Ren, Yi and Sutherland, Danica J.},
  booktitle={International Conference on Learning Representations},
  year={2025}
}

@inproceedings{rl_sgd,
title={Do We Need {Adam}? {Surprisingly} Strong and Sparse Reinforcement Learning with {SGD} in {LLM}s},
author={Sagnik Mukherjee and Lifan Yuan and Pavan Jayasinha and Dilek Hakkani-T{\"u}r and Hao Peng},
booktitle={Forty-third International Conference on Machine Learning},
year={2026},
url={https://openreview.net/forum?id=z31fdV4WRu}
}

@article{grpo,
  title   = {{DeepSeekMath}: Pushing the Limits of Mathematical Reasoning in Open Language Models},
  author  = {Shao, Zhihong and Wang, Peiyi and Zhu, Qihao and Xu, Runxin and Song, Junxiao and Bi, Xiao and Zhang, Haowei and Zhang, Mingchuan and Li, Y.K. and Wu, Y. and Guo, Daya},
  journal = {arXiv preprint arXiv:2402.03300},
  year    = {2024},
  url     = {https://arxiv.org/abs/2402.03300}
}

@inproceedings{bbvi,
  title     = {Black Box Variational Inference},
  author    = {Ranganath, Rajesh and Gerrish, Sean and Blei, David M.},
  booktitle = {Artificial Intelligence and Statistics (AISTATS)},
  year      = {2014}
}

@misc{flashrl,
  author = {Liu, Liyuan and Yao, Feng and Zhang, Dinghuai and Dong, Chengyu and Shang, Jingbo and Gao, Jianfeng},
  title  = {{FlashRL}: 8Bit Rollouts, Full Power {RL}},
  year   = {2025},
  url    = {https://fengyao.notion.site/flash-rl}
}

@article{areal,
  author  = {Fu, Wei and Gao, Jiaxuan and Shen, Xujie and Zhu, Chen and Mei, Zhiyu and He, Chuyi and Xu, Shusheng and Wei, Guo and Mei, Jun and Wang, Jiashu and Yang, Tongkai and Yuan, Binhang and Wu, Yi},
  title   = {{AReaL}: A Large-Scale Asynchronous Reinforcement Learning System for Language Reasoning},
  journal = {arXiv preprint arXiv:2505.24298},
  year    = {2025},
  url     = {https://arxiv.org/abs/2505.24298}
}

@inproceedings{orca,
  author    = {Yu, Gyeong-In and Jeong, Joo Seong and Kim, Geon-Woo and Kim, Soojeong and Chun, Byung-Gon},
  title     = {Orca: A Distributed Serving System for Transformer-Based Generative Models},
  booktitle = {16th USENIX Symposium on Operating Systems Design and Implementation (OSDI 22)},
  pages     = {521--538},
  year      = {2022}
}

@article{single_rollout,
  author  = {Hou, Zhenyu and Li, Yujiang and Tang, Jie and Dong, Yuxiao},
  title   = {Single-Rollout Asynchronous Optimization for Agentic Reinforcement Learning},
  journal = {arXiv preprint arXiv:2607.07508},
  year    = {2026},
  url     = {https://arxiv.org/abs/2607.07508}
}

@article{ionides,
  author  = {Ionides, Edward L.},
  title   = {Truncated Importance Sampling},
  journal = {Journal of Computational and Graphical Statistics},
  volume  = {17},
  number  = {2},
  pages   = {295--311},
  year    = {2008},
  doi     = {10.1198/106186008X320456}
}

@book{owen_mc,
  author = {Owen, Art B.},
  title  = {Monte Carlo Theory, Methods and Examples},
  year   = {2013},
  url    = {https://artowen.su.domains/mc/}
}

@article{distillation,
  author  = {Hinton, Geoffrey and Vinyals, Oriol and Dean, Jeff},
  title   = {Distilling the Knowledge in a Neural Network},
  journal = {arXiv preprint arXiv:1503.02531},
  year    = {2015},
  url     = {https://arxiv.org/abs/1503.02531}
}

@article{deepseek_r1,
  author  = {{DeepSeek-AI}},
  title   = {{DeepSeek-R1}: Incentivizing Reasoning Capability in {LLM}s via Reinforcement Learning},
  journal = {arXiv preprint arXiv:2501.12948},
  year    = {2025},
  url     = {https://arxiv.org/abs/2501.12948}
}

@article{greensmith,
  author  = {Greensmith, Evan and Bartlett, Peter L. and Baxter, Jonathan},
  title   = {Variance Reduction Techniques for Gradient Estimates in Reinforcement Learning},
  journal = {Journal of Machine Learning Research},
  volume  = {5},
  pages   = {1471--1530},
  year    = {2004}
}

@article{dr_grpo,
  author  = {Liu, Zichen and Chen, Changyu and Li, Wenjun and Qi, Penghui and Pang, Tianyu and Du, Chao and Lee, Wee Sun and Lin, Min},
  title   = {Understanding {R1-Zero}-Like Training: A Critical Perspective},
  journal = {arXiv preprint arXiv:2503.20783},
  year    = {2025},
  url     = {https://arxiv.org/abs/2503.20783}
}

@book{bishop,
  author    = {Bishop, Christopher M.},
  title     = {Pattern Recognition and Machine Learning},
  publisher = {Springer},
  year      = {2006}
}

@article{qwen3,
  author  = {Yang, An and Li, Anfeng and Yang, Baosong and Zhang, Beichen and Hui, Binyuan and Zheng, Bo and Yu, Bowen and Gao, Chang and others},
  title   = {Qwen3 Technical Report},
  journal = {arXiv preprint arXiv:2505.09388},
  year    = {2025},
  url     = {https://arxiv.org/abs/2505.09388}
}

@article{stream_of_search,
  author  = {Gandhi, Kanishk and Lee, Denise and Grand, Gabriel and Liu, Muxin and Cheng, Winson and Sharma, Archit and Goodman, Noah D.},
  title   = {Stream of Search ({SoS}): Learning to Search in Language},
  journal = {arXiv preprint arXiv:2404.03683},
  year    = {2024},
  url     = {https://arxiv.org/abs/2404.03683}
}

@article{intellect2,
  author  = {{Prime Intellect Team} and Jaghouar, Sami and Mattern, Justus and Ong, Jack Min and Straube, Jannik and Basra, Manveer and Pazdera, Aaron and Thaman, Kushal and Di Ferrante, Matthew and Gabriel, Felix and Obeid, Fares and Erdem, Kemal and Keiblinger, Michael and Hagemann, Johannes},
  title   = {{INTELLECT-2}: A Reasoning Model Trained Through Globally Decentralized Reinforcement Learning},
  journal = {arXiv preprint arXiv:2505.07291},
  year    = {2025},
  url     = {https://arxiv.org/abs/2505.07291}
}

@article{jet_rl,
  title={{Jet-RL}: Enabling On-Policy {FP8} Reinforcement Learning with Unified Training and Rollout Precision Flow},
  author={Xi, Haocheng and Ruan, Charlie and Liao, Peiyuan and Lin, Yujun and Cai, Han and Zhao, Yilong and Yang, Shuo and Keutzer, Kurt and Han, Song and Zhu, Ligeng},
  journal={arXiv preprint arXiv:2601.14243},
  year={2026}
}

@inproceedings{probing,
title={Probing {RLVR} Training Instability through the Lens of Objective-Level Hacking},
author={Yiming Dong and Kun Fu and Haoyu Li and Xinyuan Zhu and Yurou Liu and Lijing Shao and Jieping Ye and Zheng Wang},
booktitle={Forty-third International Conference on Machine Learning},
year={2026},
url={https://openreview.net/forum?id=KlGj06E8Wa}
}

@article{r3,
  title={Stabilizing {MoE} Reinforcement Learning by Aligning Training and Inference Routers},
  author={Ma, Wenhan and Zhang, Hailin and Zhao, Liang and Song, Yifan and Wang, Yudong and Sui, Zhifang and Luo, Fuli},
  journal={arXiv preprint arXiv:2510.11370},
  year={2025}
}

@article{fp8_rl,
  title={{FP8-RL}: A Practical and Stable Low-Precision Stack for {LLM} Reinforcement Learning},
  author={Qiu, Zhaopeng and Yu, Shuang and Zhang, Jingqi and Zhang, Shuai and Huang, Xue and Yang, Jingyi and Lai, Junjie},
  journal={arXiv preprint arXiv:2601.18150},
  year={2026},
  url={https://arxiv.org/abs/2601.18150}
}

@article{fp16_rl,
  title={Defeating the Training-Inference Mismatch via {FP16}},
  author={Qi, Penghui and Liu, Zichen and Zhou, Xiangxin and Pang, Tianyu and Du, Chao and Lee, Wee Sun and Lin, Min},
  journal={arXiv preprint arXiv:2510.26788},
  year={2025},
  url={https://arxiv.org/abs/2510.26788}
}

@article{negative_reinforcement,
  title={The Surprising Effectiveness of Negative Reinforcement in {LLM} Reasoning},
  author={Zhu, Xinyu and Xia, Mengzhou and Wei, Zhepei and Chen, Wei-Lin and Chen, Danqi and Meng, Yu},
  journal={arXiv preprint arXiv:2506.01347},
  year={2025},
  url={https://arxiv.org/abs/2506.01347}
}

@inproceedings{nthr,
  title={On the Effect of Negative Gradient in Group Relative Deep Reinforcement Optimization},
  author={Deng, Wenlong and Ren, Yi and Li, Muchen and Sutherland, Danica J. and Li, Xiaoxiao and Thrampoulidis, Christos},
  booktitle={Advances in Neural Information Processing Systems},
  year={2025},
  url={https://arxiv.org/abs/2505.18830}
}

@article{alp,
  title={Adaptive Layerwise Perturbation: Unifying Off-Policy Corrections for {LLM} {RL}},
  author={Ye, Chenlu and Zhang, Xuanchang and Hao, Yifan and Yu, Zhou and Zhang, Ziji and Gullapalli, Abhinav and Chen, Hao and Huang, Jing and Zhang, Tong},
  journal={arXiv preprint arXiv:2603.19470},
  year={2026},
  url={https://arxiv.org/abs/2603.19470}
}

@online{flashinfer_sentinel,
  title = {fix(attention): handle extreme negative logits in masked softmax},
  author = {{FlashInfer contributors}},
  year = {2026},
  month = aug,
  url = {https://github.com/flashinfer-ai/flashinfer/pull/4401}
}

@misc{o1,
      title={{OpenAI} {o1} System Card}, 
      author={{OpenAI}},
      year={2024},
      eprint={2412.16720},
      archivePrefix={arXiv},
      primaryClass={cs.AI},
      url={https://arxiv.org/abs/2412.16720}, 
}
\bibliographystyle{iclr2027_conference}

\appendix
\clearpage

\section{Implementing Score Centering}
\label{sec:sc_implementation}

\begin{notationbox}
\textbf{Notation} (as in \Cref{sec:sc}). At a fixed prefix, $y_t$ is the sampled token and $v$ a vocabulary token; $p_v$, $q_v$ are its trainer and sampler probabilities, $s_v = \nabla_\theta \log p_v$ its score, and $\bar{s} = \sum_v q_v \, s_v$ the expected score under the sampler.
\end{notationbox}

\subsection{Top-$k$ Approximation}

\textit{Full} score centering, as we have described it thus far, is prohibitively expensive for most practical applications. Computing the score centering correction term (an expectation of scores over the sampler's per-token distribution) as described in \Cref{eq:sc} requires storing the full output logprobs for every sampled token. For instance, Qwen3 models have a vocabulary size of 152K, so storing the full logprobs for just a single token in fp32 precision requires around 608KB of memory. Assuming a batch of 1024 rollouts, each with a sequence length of 32K, storing the full logprobs would require a prohibitive 20TB of memory. For this reason, we only store a top-$k$ approximation of the sampler's distribution throughout our experiments, where $k=128$. However, rather than merely computing the expected score over the top-$k$ distribution, we still try to approximate the sampler's full distribution, using logprobs from the trainer to fill in the tail. We used $k=128$ as a conservative default value; \Cref{fig:top_k} shows that $k=128$ and even $k=32$ perform on par with full score centering across all of our experimental settings.

The question is then how to reconstruct the sampler's full distribution from only its top-$k$ log-probabilities. 
Let $\hat{q}$ denote our approximation of the sampling distribution. 
We take top-$k$ log-probabilities from the sampler and model the tail using logprobs from the trainer, rescaled so that the entire distribution sums to one. Denoting the set of top-$k$ tokens as head $H$ and the remaining tokens as tail $T$:
\begin{equation}
\hat{q}_v =
\begin{cases}
q_v & v \in H \\[2pt]
\rho \, p_v & v \in T
\end{cases}
\qquad \textrm{where} \quad
\rho = \frac{1 - \sum_{v \in H} q_v}{1 - \sum_{v \in H} p_v}.
\label{eq:tail_model}
\end{equation}
The scalar $\rho$ is the ratio of the sampler's tail mass to the trainer's tail mass -- it rescales the trainer's tail so that $\hat{q}$ sums to one.

To compute the expected score under this reconstructed $\hat{q}$ distribution, there is no need to sum over the whole vocabulary. Instead, we compute the expectation over the head taken from $q$ exactly, and compute the expectation over the tail taken from $p$ as zero minus the expectation over the head, taking advantage of the fact that the expected score over the full vocabulary sums to zero:
\begin{equation}
\mathbb{E}_{\hat{q}}[s_{y_t}]
= \sum_{v \in H} q_v \, s_v + \rho \sum_{v \in T} p_v \, s_v
= \sum_{v \in H} q_v \, s_v + \rho \, \Big( \underbrace{\mathbb{E}_p[s_{y_t}]}_{=\,0} - \sum_{v \in H} p_v \, s_v \Big)
= \sum_{v \in H} \left( q_v - \rho \, p_v \right) s_v.
\label{eq:sc_topk}
\end{equation}
As a result, the centering term only involves the $k$ head tokens, even though the modeled tail covers the whole vocabulary.

The cost of top-$k$ score centering is negligible: on matched hardware, runs with $k=128$ finish within 1\% of the wall-clock time of the baseline methods for both the 0.6B and 30B models. Our custom JAX sampler computes the top-$k$ logprobs as part of decoding; vLLM and SGLang expose top-$k$ logprobs natively, but we have not measured their overhead.

\subsection{Composing Score Centering with Importance Sampling}
\label{sec:is_composition}

Score centering subtracts a correction term equal to the expected score. Importance sampling methods like TIS or MIS work by reweighting the scores using clipped / masked importance ratios. Therefore, when composing score centering on top of importance sampling, we compute the expectation of the \textit{weighted} scores rather than raw scores. We can think of this as IS first reweighting the scores, and then score centering being applied on top of the reweighted scores.

\clearpage
Let $r_v = p_v / q_v$ denote the importance ratio and $w_v = f(r_v)$ the weight assigned by the IS method being composed, for example $f(r) = \min(r, 2)$ for TIS. Score centering subtracts the expectation of the \textit{weighted} score:
\begin{equation}
R \left( w_{y_t} \, s_{y_t} - \mathbb{E}_q[w_{y_t} \, s_{y_t}] \right),
\qquad
\mathbb{E}_q[w_{y_t} \, s_{y_t}] = \sum_v q_v \, w_v \, s_v.
\label{eq:sc_is}
\end{equation}
Just like vanilla score centering, the expected update under a constant reward is exactly zero -- subtracting the expected weighted score $\mathbb{E}_q[w_{y_t} \, s_{y_t}]$ cancels drift by construction, regardless of the weighting function $f(r)$.

The top-$k$ reduction in \Cref{eq:sc_topk} also applies here. On the modeled tail the sampler probability is only known through $\hat{q}$, so the weight is computed under $\hat{q}$ as well: $\hat{w}_v = f(p_v / \hat{q}_v)$, which equals $w_v$ on the head. Since $p_v / \hat{q}_v = p_v / (\rho \, p_v) = 1/\rho$ is constant on the tail, the centering term keeps the same head-only form, with a scalar $\alpha$ in place of $\rho$:
\begin{equation}
\mathbb{E}_{\hat{q}}[\hat{w}_{y_t} \, s_{y_t}] = \sum_{v \in H} \left( q_v \, w_v - \alpha \, p_v \right) s_v,
\qquad \textrm{where} \quad
\alpha = \rho \, f(1/\rho).
\label{eq:sc_is_topk}
\end{equation}
Vanilla score centering is just a special case of $f = 1$, giving $\alpha = \rho$. As a sanity check, vanilla importance sampling $f(r) = r$ gives $\alpha = 1$ and $q_v \, w_v = p_v$, so the centering term vanishes -- exact importance sampling already has zero expected weighted score, so there is nothing left to center.

\subsection{Loss Formulation}

Score centering can be efficiently implemented in autograd frameworks by expressing it inside a scalar loss function. Since the expected score is a weighted sum of per-token scores, $\bar{s} = \sum_v q_v \, \nabla_\theta \log p_v$, we can express score centering as a loss by weighting the trainer's logprobs with detached sampler probabilities. For policy gradient this becomes:
\begin{equation}
L = -R \, \Big( \log p_{y_t} - \underbrace{\sum_v \operatorname{sg}\left[ q_v \right] \log p_v}_{\text{centering term}} \Big),
\label{eq:sc_loss}
\end{equation}
where $\operatorname{sg}[\cdot]$ denotes stop-gradient, i.e.\ we do not differentiate through the sampling probabilities. Differentiating recovers the centered update: $-\nabla_\theta L = R \, (s_{y_t} - \bar{s}) = R \, \tilde{s}_{y_t}$. The generalized top-$k$ version, composed on top of arbitrary importance weights, follows the same pattern, requiring only computation over the head tokens:
\begin{equation}
L = -R \, \Big( \operatorname{sg}\left[ w_{y_t} \right] \log p_{y_t} - \sum_{v \in H} \operatorname{sg}\left[ q_v \, w_v - \alpha \, p_v \right] \log p_v \Big).
\label{eq:sc_loss_topk}
\end{equation}

The JAX code below implements \Cref{eq:sc_loss_topk}, with \texttt{weight\_fn} specifying $f(r)$ (default: vanilla SC). As in \Cref{sec:sc_impl}, \texttt{train\_logp} spans the vocabulary and \texttt{samp\_logp} contains the sampler's top-$k$ logprobs. The additional scalar \texttt{samp\_token\_logp} is the sampler's logprob for the sampled token, even if it falls outside the head. Tail masses are floored at \texttt{eps} for numerical stability.
\par\noindent
\begin{minipage}{\linewidth}
\begin{lstlisting}[style=python]
import jax.numpy as jnp
from jax.lax import stop_gradient

def score_centering_loss(train_logp, samp_logp, topk_ids, sampled_token,
                         samp_token_logp, advantage, weight_fn=lambda r: 1.0, eps=1e-6):
    train_head_logp = train_logp[topk_ids]
    head_weights = weight_fn(jnp.exp(train_head_logp - samp_logp))
    train_tail_mass = jnp.maximum(1 - jnp.exp(train_head_logp).sum(), eps)
    samp_tail_mass = jnp.maximum(1 - jnp.exp(samp_logp).sum(), eps)
    tail_mass_ratio = samp_tail_mass / train_tail_mass
    tail_scale = tail_mass_ratio * weight_fn(1 / tail_mass_ratio)
    head_prob_residual = jnp.exp(samp_logp) * head_weights
    head_prob_residual -= tail_scale * jnp.exp(train_head_logp)
    logp_correction = (stop_gradient(head_prob_residual) * train_head_logp).sum()
    sampled_ratio = jnp.exp(train_logp[sampled_token] - samp_token_logp)
    weighted_logp = stop_gradient(weight_fn(sampled_ratio)) * train_logp[sampled_token]
    return -advantage * (weighted_logp - logp_correction)

tis_weight = lambda r: jnp.minimum(r, 2.0)
mis_weight = lambda r: jnp.where((r >= 0.5) & (r <= 5.0), r, 0.0)
\end{lstlisting}
\end{minipage}
Pass \texttt{weight\_fn=tis\_weight} or \texttt{weight\_fn=mis\_weight} to compose score centering with TIS or MIS, respectively. 
As in \Cref{eq:sc_loss_topk}, both the sampled importance weight and the centering coefficients are detached.

\subsection{Top-$k$ Ablation}
\label{sec:topk_ablation}

\begin{figure}[h]
    \centering
    \includegraphics[width=1\linewidth]{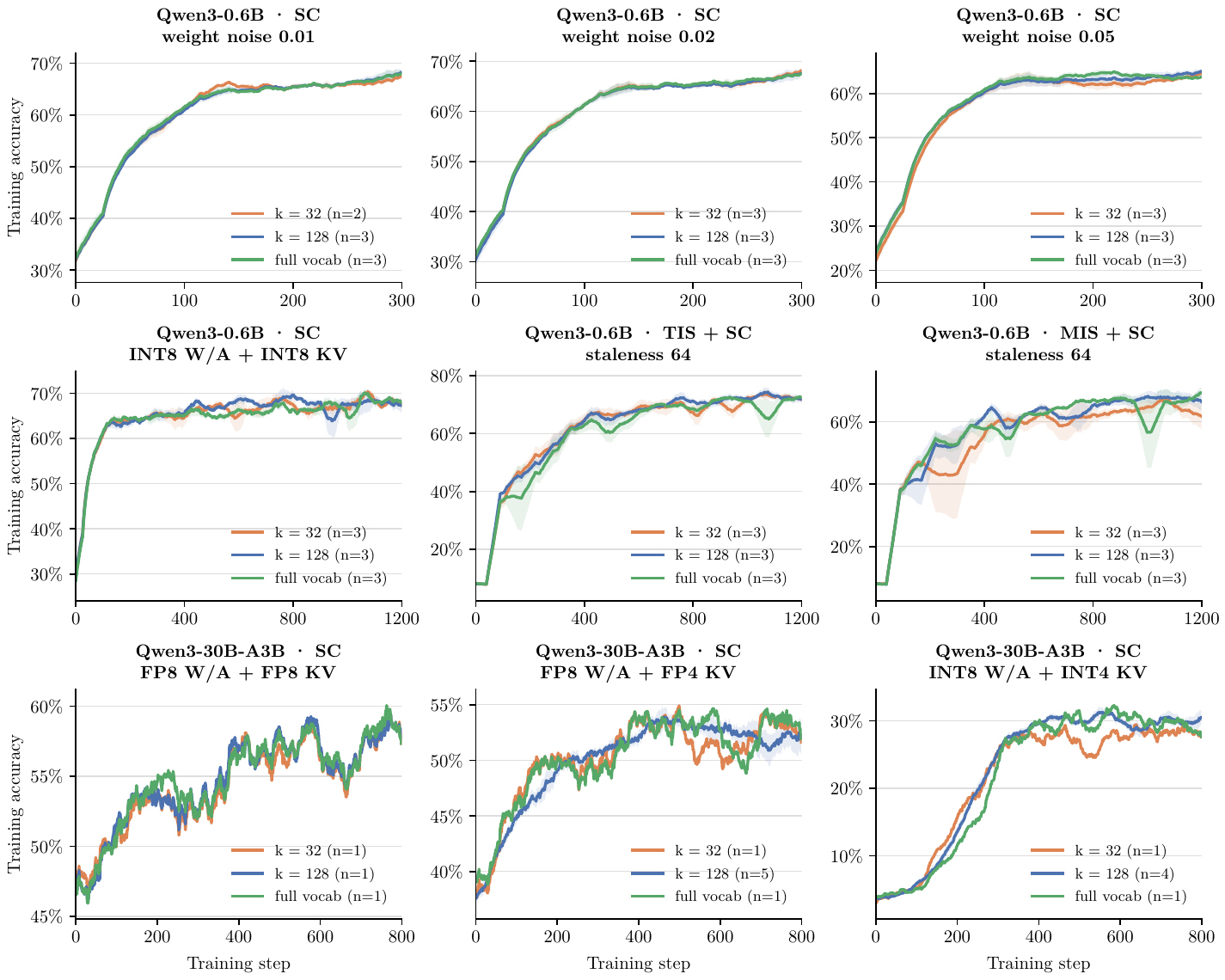}
    \caption{Top-$k$ ablation of score centering. Using $k=128$ or $k=32$ sampler logprobs with the tail model of \Cref{eq:tail_model} matches full-vocabulary score centering across all tested settings. The number of seeds for each curve is printed in the legend.}
    \label{fig:top_k}
\end{figure}

In \Cref{fig:top_k}, we compare top-$k$ score centering against full score centering across all of our experimental settings. 
Both $k=32$ and $k=128$ perform on par with full score centering in every setting. 
The tail model of \Cref{eq:tail_model} only has to account for the sampler probability mass outside the top-$k$ head, which we log at every step. With $k=128$, the head covers more than $99.9\%$ of the sampler's mass on average in every setting except INT8 W/A + INT4 KV at 30B, where it covers $99.45\%$ on average and $95.8\%$ in the worst batch -- the setting where the tail is most distorted by quantization and where \Cref{fig:top_k} shows that $k=32$ and $k=128$ still match full score centering.

\section{Experimental Details}
\label{sec:experimental_details}

\subsection{Training Setup}

Unless otherwise noted, we use REINFORCE with group-centered rewards:
\begin{equation}
J_{\mathrm{base}}(\theta)
= \frac{\sum_i A_i \sum_{t=1}^{L_i} \log p_\theta(y_{i,t}\mid y_{i,<t})}
       {\sum_i L_i},
\qquad A_i = R_i - \bar R_{\mathrm{group}},
\qquad R_i \in \{-1,+1\}.
\label{eq:shared_reinforce}
\end{equation}
Here $i$ indexes rollouts in the batch, $L_i$ is the rollout length, and $\bar R_{\mathrm{group}}$ is the mean reward of completions for the same prompt.

Following~\citet{rl_sgd}, we use SGD across all of our experiments, with a fixed learning rate of $10^{-2}$. We found this learning rate stable across all of our experiments, and performing on par with AdamW, while saving up to 240GB of memory. Each batch samples 8 completions per prompt and takes a single optimizer step. We use 64 prompts per batch (512 sequences) with a maximum sequence length of 512 for Qwen3-0.6B-Instruct on Countdown, and 16 prompts per batch (128 sequences) with a maximum sequence length of 1024 for Qwen3-30B-A3B-Base on INTELLECT-2 math. Longer sequences would increase TIM \citep{jet_rl} but make multi-seed comparison of many baselines too expensive.

\subsection{Correction Methods}
\label{sec:is_methods}

All of the IS-based methods we compare fall onto a simple grid (\Cref{tab:is_methods}). They differ in whether the correction is applied per token or per sequence, where the importance ratio gets clipped or masked, and which advantage signs are corrected. DPPO uses a masking region based on binary total variation instead of the importance ratio, but its objective still uses IS weights. We use default parameters from the original papers or \texttt{verl} without further tuning. All importance ratios are computed against the sampler's logged probabilities. GSPO uses the geometric mean of the token ratios, as in the original paper.

We apply each correction on top of the same objective, rather than reproducing the full recipe from each paper (\Cref{sec:results_setup}). The last column of \Cref{tab:is_methods} provides implementation details for each method. PG uses \Cref{eq:shared_reinforce} without any correction. For score centering, we use the top-$k$ approximation with $k=128$ (\Cref{sec:sc_implementation}).

\begin{table}[t]
    \centering
    \caption{IS-based correction methods represented as a grid, with implementation notes in the last column. \texttt{Level} determines whether the statistic is computed per token or per sequence. \texttt{Outside} specifies what happens beyond the \texttt{Low...High} region. \texttt{Sign} specifies which advantage signs are corrected: \texttt{both} applies the weight to every token; \texttt{neg} only to tokens with negative advantage (positive-advantage tokens keep weight 1); and \texttt{ppo} applies the PPO pessimistic clip, i.e.\ the minimum of the weighted and clipped objectives, so the clip only binds in the direction that would increase the objective. Dual clipping follows \texttt{verl}'s PPO implementation: negative-advantage contributions with ratio above $3$ are dropped (whole responses for GSPO). $^{\mathrm P}$ denotes the paper default and $^{\mathrm V}$ the \texttt{verl} default \citep{verl}.}
    \label{tab:is_methods}
    \small
    \arrayrulecolor{gridline}
    \setlength{\arrayrulewidth}{0.3pt}
    \setlength{\tabcolsep}{3pt}
    \renewcommand{\arraystretch}{1.3}
    \begin{tabular}{@{}>{\raggedright\arraybackslash}m{70pt}
        >{\centering\arraybackslash}m{25pt}
        >{\centering\arraybackslash}m{24pt}
        >{\centering\arraybackslash}m{30pt}
        >{\centering\arraybackslash}m{30pt}
        >{\centering\arraybackslash}m{32pt}
        >{\centering\arraybackslash}m{22pt}
        >{\raggedright\arraybackslash}m{\dimexpr\linewidth-275pt\relax}@{}}
        \hline
        \rowcolor{gridgray}
        \textbf{Method} & \textbf{Level} & \textbf{Stat} & \textbf{Low} & \textbf{High} & \textbf{Outside} & \textbf{Sign} & \textbf{Implementation notes} \\
        \hline
        \cellcolor{gridgray}Naive IS \citep{reinforce} & \cellcolor{deep0!12}\texttt{token} & \cellcolor{tblteal!12}\texttt{ratio} & \cellcolor{deep1!50}0 & \cellcolor{deep1!50}$\infty$ & \cellcolor{gentlelilac!5}\texttt{-} & \cellcolor{deep2!30}\texttt{both} & --- \\
        \hline
        \cellcolor{gridgray}TIS$^{\mathrm V}$ \citep{tis}\par CISPO \citep{cispo} & \cellcolor{deep0!12}\texttt{token} & \cellcolor{tblteal!12}\texttt{ratio} & \cellcolor{deep1!50}0 & \cellcolor{deep1!30}2.0 & \cellcolor{gentlelilac!30}\texttt{clip} & \cellcolor{deep2!30}\texttt{both} & CISPO's dynamic sampling and length penalty omitted. \\
        \hline
        \cellcolor{gridgray}MIS (IcePop)$^{\mathrm P}$ \citep{icepop} & \cellcolor{deep0!12}\texttt{token} & \cellcolor{tblteal!12}\texttt{ratio} & \cellcolor{deep1!30}0.5 & \cellcolor{deep1!30}5.0 & \cellcolor{gentlelilac!10}\texttt{drop} & \cellcolor{deep2!30}\texttt{both} & Separate PPO surrogate between policy versions omitted. \\
        \hline
        \cellcolor{gridgray}PPO$^{\mathrm{P,V}}$ \citep{ppo} & \cellcolor{deep0!12}\texttt{token} & \cellcolor{tblteal!12}\texttt{ratio} & \cellcolor{deep1!16}0.8 & \cellcolor{deep1!16}1.2 & \cellcolor{gentlelilac!10}\texttt{drop} & \cellcolor{deep2!12}\texttt{ppo} & Dual clipping at $3$ (\texttt{verl} default). \\
        \hline
        \cellcolor{gridgray}DAPO$^{\mathrm{P,V}}$ \citep{dapo} & \cellcolor{deep0!12}\texttt{token} & \cellcolor{tblteal!12}\texttt{ratio} & \cellcolor{deep1!16}0.8 & \cellcolor{deep1!16}1.28 & \cellcolor{gentlelilac!10}\texttt{drop} & \cellcolor{deep2!12}\texttt{ppo} & Dual clipping at $3$; dynamic sampling and overlong reward penalty omitted. \\
        \hline
        \cellcolor{gridgray}GSPO$^{\mathrm{P,V}}$ \citep{gspo} & \cellcolor{deep0!38}\texttt{seq} & \cellcolor{tblteal!12}\texttt{ratio} & \cellcolor{deep1!6}0.9997 & \cellcolor{deep1!6}1.0004 & \cellcolor{gentlelilac!10}\texttt{drop} & \cellcolor{deep2!12}\texttt{ppo} & Token gradients summed rather than response-averaged; added dual clipping at $3$. \\
        \hline
        \cellcolor{gridgray}TOPR$^{\mathrm P}$ \citep{topr} & \cellcolor{deep0!38}\texttt{seq} & \cellcolor{tblteal!12}\texttt{ratio} & \cellcolor{deep1!50}0 & \cellcolor{deep1!6}1.0 & \cellcolor{gentlelilac!30}\texttt{clip} & \cellcolor{deep2!50}\texttt{neg} & --- \\
        \hline
        \cellcolor{gridgray}DPPO$^{\mathrm P}$ \citep{dppo} & \cellcolor{deep0!12}\texttt{token} & \cellcolor{tblteal!38}\texttt{tv} & \cellcolor{deep1!50}0 & \cellcolor{deep1!30}0.2 & \cellcolor{gentlelilac!10}\texttt{drop} & \cellcolor{deep2!12}\texttt{ppo} & No importance-ratio clipping. \\
        \hline
    \end{tabular}
\end{table}

\subsection{Compute Resources}

Across all figures, shading represents $\pm 1$ standard error. \Cref{fig:2b_online_vs_offline,fig:06b_weight_noise} report accuracy on held-out evaluation prompts, while \Cref{fig:30b_math,fig:06b_quant_stale,fig:top_k} report training accuracy, i.e.\ the mean reward of the sampled rollouts, smoothed with a centered moving average over training steps.
\Cref{tab:reproduction_compute} lists the number of seeds per method and the approximate compute required to reproduce each figure.
For the 0.6B and 1.7B models, we ran 3 seeds across every experiment for every correction method (\Cref{fig:2b_online_vs_offline,fig:06b_weight_noise,fig:06b_quant_stale}). However, since training the 30B model (\Cref{fig:30b_math}) was much more expensive, we initially ran every method only with a single seed, and then added more seeds only for those methods that performed best (since these were the comparisons we cared about the most). In the FP8 KV panel every method is single-seed, while in the FP4 KV and INT4 KV panels score centering uses 5 and 4 seeds, TIS 4 and 3, MIS 3 and 3, and vanilla IS 2 and 1, respectively. 
All remaining methods are single-seed. Similarly, in \Cref{fig:top_k}, the $k=128$ results for the 30B model are transferred from \Cref{fig:30b_math}, which is why they are multi-seed; $k=32$ and full score centering are single-seed to reduce computational cost.

\begin{table}[H]
    \centering
    \caption{Seeds per curve and approximate compute required to reproduce our figures. Experiments primarily ran on nodes with 8$\times$ NVIDIA H100 SXM GPUs.}
    \label{tab:reproduction_compute}
    \resizebox{\linewidth}{!}{%
    \arrayrulecolor{gridline}
    \setlength{\arrayrulewidth}{0.3pt}
    \begin{tabular}{lllrr}
        \hline
        \rowcolor{gridgray}
        \textbf{Experiment} & \textbf{Figure} & \textbf{Seeds} & \textbf{Unique runs} & \textbf{Approx. H100-hours} \\
        \hline
        \cellcolor{gridgray}Online vs.\ offline & \Cref{fig:2b_online_vs_offline} & 3 & 18 & 190 \\
        \hline
        \cellcolor{gridgray}Synthetic weight noise & \Cref{fig:06b_weight_noise} & 3 & 108 & 280 \\
        \hline
        \cellcolor{gridgray}Quantization and staleness & \Cref{fig:06b_quant_stale} & 3 & 72 & 1,120 \\
        \hline
        \cellcolor{gridgray}30B math & \Cref{fig:30b_math} & 1--5 & 47 & 3,750 \\
        \hline
        \cellcolor{gridgray}Top-$k$ ablation & \Cref{fig:top_k} & see legend & 51 & 840 \\
        \hline
        \cellcolor{gridgray}\textbf{Total} & & & \textbf{296} & \textbf{6,180} \\
        \hline
    \end{tabular}%
    }
\end{table}

\end{document}